%% file: sample-sigconf.tex
\documentclass[sigconf]{acmart}
\AtBeginDocument{%
  \providecommand\BibTeX{{%
    \normalfont B\kern-0.5em{\scshape i\kern-0.25em b}\kern-0.8em\TeX}}}

\usepackage{graphicx}
\usepackage{subcaption}

\usepackage{booktabs}
\usepackage{bbm}
\usepackage{pifont}
\usepackage{multirow}
\usepackage{enumitem}
\usepackage[normalem]{ulem}
\usepackage{physics,amsmath}
 
\usepackage{amssymb}
\usepackage{colortbl}

\newcommand{\lyc}[1]{{\color{black}{#1}}}

\definecolor{gray}{rgb}{ 0.851,  0.851,  0.851}

\newcommand{\ie}{\emph{i.e., }}
\newcommand{\eg}{\emph{e.g., }}

\newcommand{\cf}{\emph{cf. }}
\newcommand{\aka}{\emph{a.k.a. }}

\newcommand{\blue}[1]{{\color{blue}{#1}}}

\usepackage{tikz}

\usepackage{cleveref}
\crefname{section}{Sec.}{Secs.}
\Crefname{section}{Section}{Sections}
\Crefname{table}{Table}{Tables}
\crefname{table}{Tab.}{Tabs.}

\copyrightyear{2023}
\acmYear{2023}
\setcopyright{acmlicensed}\acmConference[MM '23]{Proceedings of the 31st
ACM International Conference on Multimedia}{October 29-November 3,
2023}{Ottawa, ON, Canada}
\acmBooktitle{Proceedings of the 31st ACM International Conference on
Multimedia (MM '23), October 29-November 3, 2023, Ottawa, ON, Canada}
\acmPrice{15.00}
\acmDOI{10.1145/3581783.3612577}
\acmISBN{979-8-4007-0108-5/23/10}

\begin{document}

\title{Redundancy-aware Transformer for Video Question Answering}

% \author{Yicong Li$^1$, Junbin Xiao$^1$, Chun Feng$^2$, Xiang Wang$^2$, Tat-Seng Chua$^1$\\
% $^1$National University of Singapore, 
% $^2$University of Science and Technology of China,\\
% {\tt\small liyicong@u.nus.edu,fengchun3364@mail.ustc.edu.cn, xiangwang1223@gmail.com} \\
% {\tt\small junbin@comp.nus.edu.sg,  dcscts@nus.edu.sg}
% }.

\author{Yicong Li$^{1}$, Xun Yang$^{2*}$, An Zhang$^{1}$, Chun Feng$^{2}$, Xiang Wang$^{2}$, Tat-Seng Chua$^1$}
\def\authors{Yicong Li, Xun Yang, An Zhang, Chun Feng, Xiang Wang, and Tat-Seng Chua}
\affiliation{
\institution{$^1$National University of Singapore, $^2$University of Science and Technology of China}
\country{}
}
\email{liyicong@u.nus.edu,{hfutyangxun,xiangwang1223}@gmail.com}
\email{fengchun3364@mail.ustc.edu.cn, {an_zhang,dcscts}@nus.edu.sg}
\thanks{$*$ Xun Yang is the corresponding author}
\thanks{Xiang Wang is also affiliated with Institute of Artificial Intelligence, Institute of Dataspace, Hefei Comprehensive National Science Center.}

\renewcommand{\shortauthors}{Li et al.}

% \renewcommand{\shortauthors}{Li et al.}

\begin{abstract}
    This paper identifies two kinds of redundancy in the current VideoQA paradigm. Specifically, the current video encoders tend to holistically embed all video clues at different granularities in a hierarchical manner, which inevitably introduces \textit{neighboring-frame redundancy} that can overwhelm detailed visual clues at the object level. Subsequently, prevailing vision-language fusion designs introduce the \textit{cross-modal redundancy} by exhaustively fusing all visual elements with question tokens without explicitly differentiating their pairwise vision-language interactions, thus making a pernicious impact on the answering.
    
    To this end, we propose a novel transformer-based architecture, that aims to model VideoQA in a redundancy-aware manner. To address the neighboring-frame redundancy, we introduce a video encoder structure that emphasizes the object-level change in neighboring frames, while adopting an out-of-neighboring message-passing scheme that imposes attention only on distant frames. As for the cross-modal redundancy, we equip our fusion module with a novel adaptive sampling, which explicitly differentiates the vision-language interactions by identifying a small subset of visual elements that exclusively support the answer. Upon these advancements, we find this \underline{R}edundancy-\underline{a}ware trans\underline{former} (RaFormer) can achieve state-of-the-art results on multiple VideoQA benchmarks. 
    % Code is available at  \url{https://github.com/yl3800/RaFormer}.
\end{abstract}
\vspace{-15pt}



\vspace{-15pt}

\maketitle

\input{sec/1_intro}

\input{sec/5_related}

\input{sec/2_preliminaries}

\input{sec/3_method}

\input{sec/4_results}

\input{sec/6_conclusions}

\section{Acknowledgments}
This work was supported by NExT search center, the National Natural Science Foundation of China (NSFC) under Grant 9227010114, Grant62272435,
Grant U22A2094, and the University Synergy Innovation Program of Anhui Province (GXXT-2022-040).
\clearpage

% \begin{acks}
% \end{acks}

\bibliographystyle{ACM-Reference-Format}
\bibliography{sample-base}

% %%
% %% If your work has an appendix, this is the place to put it.
% \appendix



% Lorem ipsum dolor sit amet, consectetur adipiscing elit. 


% Etiam commodo feugiat nisl pulvinar pellentesque. Etiam 


% Nam id fermentum dui. Suspendisse sagittis tortor a nulla 

\end{document}

%% file: sec/1_intro.tex
\section{Introduction} \label{sec:intro}

\begin{figure}
  \centering
  
  \begin{subfigure}{0.495\linewidth}
  \includegraphics[width=\linewidth]{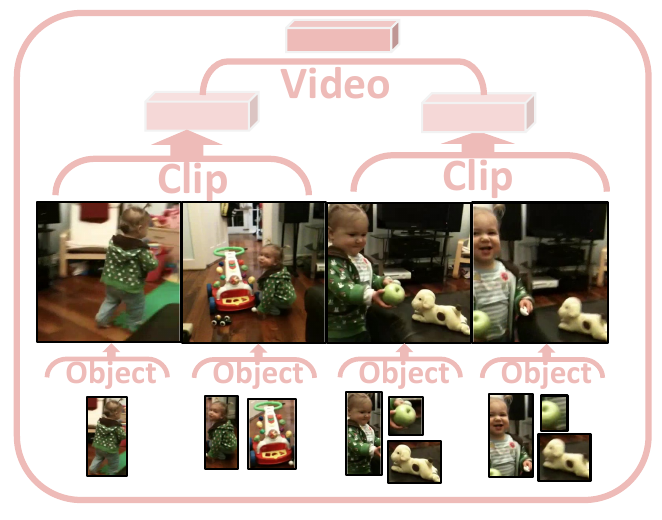}
  \caption{}
  \label{fig:1a} 
\end{subfigure}
\begin{subfigure}{0.485\linewidth}
  \includegraphics[width=\linewidth]{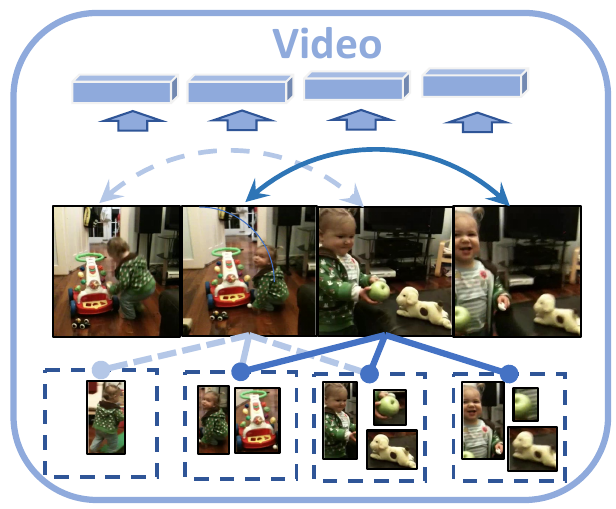}
  \caption{}
  \label{fig:1b}
\end{subfigure}

\begin{subfigure}{\linewidth}
    \centering
    \includegraphics[width=0.9\linewidth]{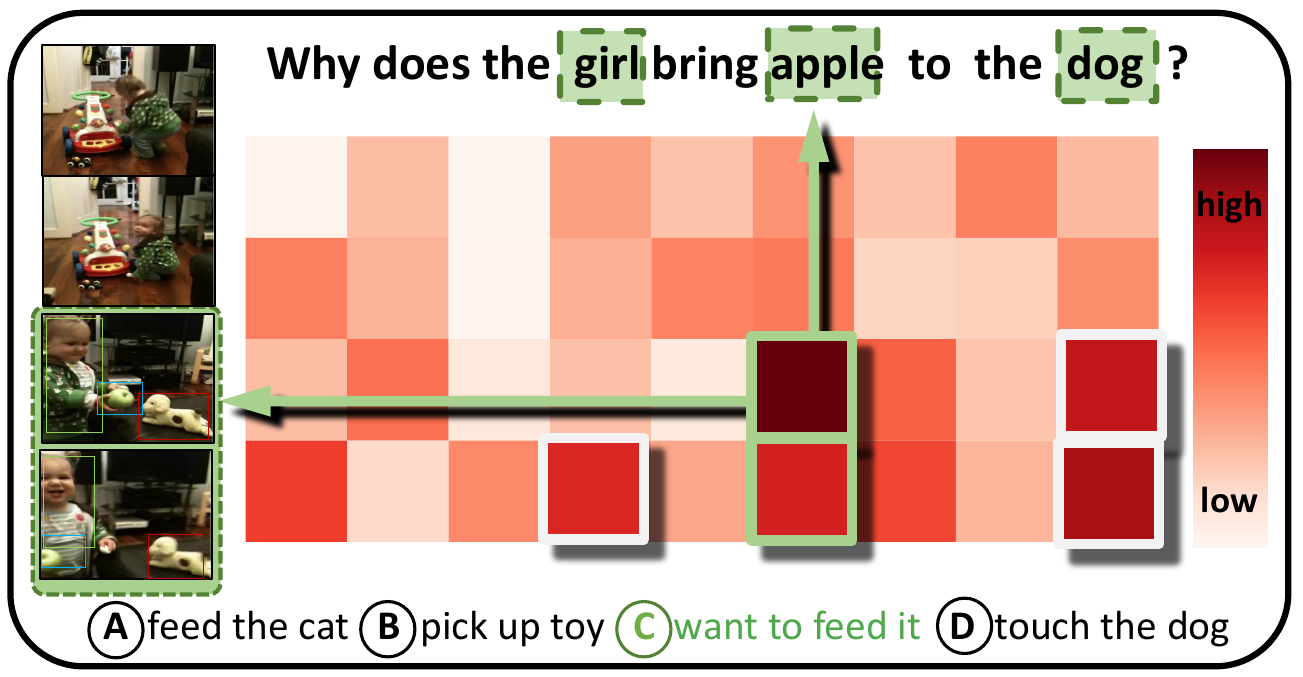}\par
    \caption{}
    \label{fig:1c} 
\end{subfigure}
% \vspace{-20pt}
\caption{(a) Conventional multi-hierarchy video encoder. (b) Our design philosophy for video encoder.
(c) An illustration of how a small proportion of interactions responds to answering, 5 critical interactions are highlighted.}
\label{fig:1}
\vspace{-10pt}
\end{figure}

Ever since the inception of intelligent systems, persistent interest has been paid to revealing how machine reflects the physical world. In this regard, Video Question Answering (VideoQA), a task that answers the natural language questions in the context of videos,  has gained steady progress over the last few years.
This advancement stems, in part from the recent surge of pre-trained foundation models \cite{VGT}, and in part from the enthusiasm for designing task-specific architectures \cite{park2021bridge,IGV,hqga}.

Scrutinizing the recent progress \cite{hga,hcrn,park2021bridge,IGV,hqga} in VideoQA, we systematize a dominant paradigm as a composition of three modules: 
1) a \textit{video encoder} that embeds the cross-frame dynamic on top of the frame representations, 
2) a \textit{question encoder} that contextualizes the question semantic to acquire text embeddings, and
3) a \textit{across-modal fuser} that encapsulates the video and question embedding to the multi-modal representations, and then makes the prediction on top of that. 
Clearly, a correct prediction requires a careful processing of visual clues and coordinating the interactions of vision-language elements to reflect the answer reasoning as a mutual agreement between video and question.

% \begin{figure}[t]
% 	\centering
% 	\subcaptionbox{\label{fig:running-example}}{
% 	   % \vspace{-5pt}
% 		\includegraphics[width=0.92\linewidth]{fig/running-example.pdf}}
% 	\subcaptionbox{ \label{fig:inductive-bias}}{
% 	   % \vspace{-5pt}
% 		\includegraphics[width=0.92\linewidth]{fig/inductive-bias.pdf}}
% 	% \vspace{-5pt}
% 	\caption{Running example. (a) An illustration of how critical interactions respond to answering, top 5 critical interactions are highlighted. (b) The pipeline of the current encoder}
% 	\label{fig:intro-example}
% 	% \vspace{-15pt}
% \end{figure}

%%%%%%%%%%%%%%%%%%%%%%%%%%%%%%%%%%%%%%%%%%%%%%%%%%%%%%%%%%%%%%%% 现有scheme 的问题
% TODO: The visual-linguistic interactions play a crucial role in VideoQA. However, the encoder-decoder paradigm models blindly. deficient in self-rationalization. conservatively. parts of interactions are deceptive . deprive. in part because. vision-language interaction
%

Although existing works have extensively adopted this paradigm, we argue that the potential of current practices is seriously undermined due to their inability in handling the substantial redundancy of VideoQA. Specifically, the redundancy comes from two aspects:

\begin{itemize}[leftmargin=*] 
    \item 
    % Prevailing \textbf{Video Encoder}  \cite{VGT, hostr, hcrn} typically modeling the input video as as a multi-level hierarchy where visual feature at different granularities are merged together in a bottom up manner to get a video-level representation.  (See \cref Fig1a) 
    % However, videos are temporally redundant, the neighboring frames are similar in general because of the shared environmental background, despite being different in micro details (\ie few foreground objects). Due to such similarity, recklessly merging all adjacent frames together is neither effective nor necessary, which introduces the \textit{neighboring-frame redundancy} that might overwhelms the detail movement of foreground objects.
    % Thus, we argue and an effective modelling of video should 1) focus on the change of objects in adjacent frames, instead of merge all visual element in each frame together. 2) impose frame-level merging only for distant frames, where the gradually change of objects has accumulated into a qualitative change at frame level.

    The prevailing \textbf{video encoder} \cite{VGT, hostr, hcrn}, typically models the input video as a multi-level hierarchy where visual features at different granularities are merged together in a bottom-up manner (as shown in \cref{fig:1a}). 
    However, videos are temporally redundant, the neighboring frames are similar in general because they share identical environment or background, despite the minor differences in foreground objects. Therefore, recklessly merging all adjacent frames together is neither effective nor necessary, which introduces the \textit{neighboring-frame redundancy} that will overwhelms the detail movement of foreground objects in adjacent frames.
    Thus, as shown in \cref{fig:1b}, we argue that an effective video modeling approach should focus on the object-level changes in adjacent frames rather than merging all visual elements in each frame together. Moreover, frame-level merging should only be imposed on distant frames, where the gradual changes of objects have accumulated into a qualitative change at the frame level.

    % \item Existing \textbf{Cross-modal Fuser} 
    % \cite{jang2017tgif,hcrn,hostr,pgat} perform holistic encoding exhaustively without modeling the pairwise interactions between video and language tokens, which makes differentiating interaction infeasible.
    % In fact, various interactions contribute differently, and only a small proportion of critical interactions are responsive to answering, leaving the rest as \textit{cross-modal redundancy}; this makes reasoning of VideoQA suffer a highly sparse target signal.
    % In fact, various interactions contribute differently, and only a small proportion of critical interactions are responsive to answering, leaving the rest as redundancy; this makes reasoning of VideoQA suffer a highly sparse target signal. Taking \cref{fig:running-example} as an example, to answer the question ``Why does the girl bring apple to the dog?'',
    % the interactions between the query words ``girl, apple, and dog'' and the ``feeding'' scene in last two clips hold the causal information that best supports the answer, while the interactions involving other clips present the environmental information only, which if not ruled out, might derail the reasoning \cite{DBLP:conf/ijcai/RossHD17,CSS,IGV}.

    \item The existing \textbf{cross-modal fuser} \cite{jang2017tgif,hcrn,hostr,pgat} performs exhaustive feature fusing, that is, merging all information from the video and language inputs, without differentiating the pairwise interactions between them. However, various interactions contribute differently, and only a small proportion of critical interactions are responsive to answering the question, leaving the rest bulk as \textit{cross-modal redundancy}. This redundancy lead to a highly sparse target signal in VideoQA, making the answer reasoning extremely challenging. 
    % Therefore, we argure the necessity important to consider that different interactions contribute differently, and only a small number of critical interactions are necessary to answer a question. 
    Taking \cref{fig:1c} as an example, the interactions between the query words "girl, apple, and dog" and the "feeding" scene in the last two clips contain the causal information that best supports the answer, while interactions involving other clips contain only environmental information and should be ruled out to avoid derailing the reasoning process \cite{DBLP:conf/ijcai/RossHD17,CSS,IGV}.
    
\end{itemize}
% this paradigm can not effectively modal the video  
% However, we argue current practices of this paradigm are deficient in handling the large volume of redundancy information in VideoQA. Specifically, their  redundancy comes from two aspects:
% \begin{itemize}[leftmargin=*]
%     \item \textbf{Prevailing Video Encoders are ineffective in modeling the visual clue at different levels.} Existing \textbf{Video Encoder} typically leverage frame-level (\eg ResNet output) and object-level (\eg ROI feature from Faster-RCNN) as the visual input,   temporal dynamic fin incorporates both frame-level (\eg ResNet output) and object-level (\eg ROI feature from Faster-RCNN) for a fine-grained inspection of visual clues. Specifically, they tend to model the video as a multi-level hierarchy, where low level visual objects are aggregated and add to the by either tend to model tend to enhance the frame representation by 
% \end{itemize}

% We argue that such non-discriminative encoding will inevitably tarnish the label information by fusing a large portion of redundancy, thus, disturbing the multi-model representation learning.

To resolve these limitations, we propose a novel VideoQA framework, named \underline{R}edundancy-\underline{a}ware Trans\underline{former} (\textbf{RaFormer}) that introduces two fundamental advancements: 1) a novel video encoder that emphasizes the object-level change within a temporal neighborhood by modelling object movement in adjacent frames, while avoid neighboring-frame redundancy by imposing attention only on distant frames, and 2) an adaptive sampling module that enables the cross-modal fuser  to discover the answer critical frames in an interaction-aware manner.
Specifically, our video encoder aims to model the detailed change of objects in adjacent frames, while imposing out-of-neighborhood message passing at frame-level to avoid redundancy.
To achieve this, we first introduce a window cross-attention that enhances a single frame representation using objects within its temporal window. Then, on top of all enhanced frame representations, we adopt leap attention \cite{leapattention} to impose connection only between two temporal distant frames, which avoid the neighboring-frame redundancy in a dilate manner. Notably, unlike conventional hierarchy-based methods \cite{VGT, hostr, hcrn} that merge visual elements in different granularities into a single packed representation (see \cref{fig:1a}), our video encoder uses a series of object enhanced frames to represent a video, which coordinates our fuser design. 
In cross-modal fusing, we first adopt a transformer-style encoder to exert all pairwise interactions. Then, based on encoder's cross-attention map, we design an adaptive sampling module to pinpoint a small set of critical interactions, which enables their corresponding frame being collected in an adaptive manner. Upon these collected critical frames, we can naturally infer the answer. Intuitively, such a frame down-sampling scheme acts as an information bottleneck \cite{infomation_bottleneck} that condenses the critical information for answering and makes it easier for a VideoQA model to capture the learning pattern, thus leading to a significant improvement over the original input.

%
% We show that this architecture instantiating our new scheme, which we name \underline{Ra}tionale-empowered Trans\underline{former} (\textbf{RaFormer}), outperforms the current state-of-the-art models across several popular benchmarks.

Our contributions are summarized as follows:
\begin{itemize}[leftmargin=*]
    \item \lyc{We address the long-ignored redundancy issue in VideoQA, which is elicited by the neighboring-frame redundancy in current practice of video encoder and the cross-modal redundancy of fuser.}
    \item We propose RaFormer, a fully transformer-based VideoQA model that avoids neighboring-frame redundancy by highlighting object-level change in adjacent frames and  the out-of-neighborhood message passing at frame-level. In addition, it also handles the cross-modal redundancy via a novel adaptive sampling module.
    \item We perform extensive experiments on four benchmark datasets to demonstrate the effectiveness of RaFormer. The results show significant improvements over previous arts. (The absolute improvements \textit{w.r.t.} accuracy are: NExT-QA \cite{next-qa} +3.5\%, CausalVid-QA\cite{causalvid} +3.9\%, MSVD-QA \cite{DBLP:conf/mm/XuZX0Z0Z17} +2.3\%, MSRVTT\cite{DBLP:conf/mm/XuZX0Z0Z17} +2.6\%.)
\end{itemize}

%% file: sec/5_related.tex
\section{Related works}
\label{sec:related}
\noindent\textbf{Video Question Answering (VideoQA).}
Video Question Answering (VideoQA) is a fundamental extension of ImageQA that incorporates a temporal component. Typically, previous efforts either leverage frame-level features as video representation or incorporated region features as an additional input, which brings extra benefit for detail-oriented question. Regardless of the input representation, current designs tend to merge all visual elements together as the video-level representation, using some inductive biases. \cite{hga} and \cite{park2021bridge} pioneered graph-based structures for video modeling based on the heterogeneity of input modality, while \cite{VGT} enabled relation reasoning via dynamic object graphs. In recent years, a prevailing line of research tend to extract a multi-level hierarchy from the video sequence. \cite{hcrn} built a bottom-up pathway by assembling information from clip-level to frame-level, while subsequent works such as \cite{hostr} and \cite{hqga} extended their approaches with an additional step in the object region. Despite these advances, existing methods overlook the redundant nature of VideoQA tasks. In contrast, we developed RaFormer to emphasize the two types of redundancy and mitigate their negative impact on answer reasoning.

% \vspace{5pt}
\noindent\textbf{Redundancy in Video.}
Video related tasks~\cite{dong2022partially,yang2022video,tan2021selective,yang2021deconfounded,yang2020weakly,dong2021dual, ivrd, ji2021vidvrd,ji2023binary, ji2023partial,videoQA_survey,DBLP:conf/mm/ShangLXJC21} have seen significant advancements in recent years, primarily driven by the adoption of 3D CNNs \cite{3dcnn} or transformer architecture \cite{attention,li2023transformer}.  While these models have shown promising results on different video tasks (including video understanding~\cite{shang2019annotating,xiao2020visual,DBLP:conf/mm/LiYSC21} and video-language tasks), there is a growing interest in developing more efficient techniques to maintain or even improve the performance, due to the redundancy nature of video. In video understanding domain, previous works have explored various approaches \cite{eff1,eff2,leapattention} to reduce the computational cost and redundancy, such as using hybrid 2D-3D architectures, group convolution, or selecting salient clips. 
For the video-language tasks, \cite{atp} is the first work that emphasizes the ``single-frame redundancy '' issue in video. However, its a analytical work that aims to provide suggestion for a dataset construction view. To the best od our knowledge, none of existing work focus on how ruling out redundancy in video can bring extra benefit to the VideoQA. 
Intuitively, a concise representaion with limited redundancy makes model easier to capture the causal pattern, thus outperforms the model trained with original inputs \cite{infomation_bottleneck}.

% \vspace{5pt}
\noindent\textbf{Rationale Discovery.}
% In pursuit of  explainability, the recent development of DNN is encouraged to reveal the intuitive evidence of their prediction, \ie the rationales. 
% As one of the prevailing practices, the rationale discovery has been extended from the NLP community\cite{DBLP:conf/kdd/Ribeiro0G16} to the Graph \cite{DIR} and Vision field \cite{DBLP:conf/cvpr/ZhangYMW19}. 
% Recently, this development also stems from the multi-modal community. 
% \cite{DBLP:conf/cvpr/ParkHARSDR18} and \cite{DBLP:conf/cvpr/DuaKB21} proposes ImageQA-based tasks that inquire about additional textual evidence, \cite{causalvid} brings this idea to the videoQA.
% Despite the progress, the recent solution either requires a rationale finder with heavy computation overhead \cite{IGV} or needs to be trained in a contrastive manner \cite{EIGV}, and they only focus on the rationale in the visual aspect. RaFormer, however, identifies cross-modal rationales without additional training loss. 
% Notably, some might categorize our implementation as a token-reduction method in the efficient transformer field \cite{DBLP:journals/corr/abs-2106-12620, dynamicViT, ats}. However, the key distinction is that they focus on the efficiency of single-modal architecture even with a trade-off of accuracy. Whereas our method discovers the rationales as the mutual agreements between modalities that bring us a consistent performance gain.
The pursuit of explainability has led to recent developments in deep neural networks (DNN) that aim to reveal the intuitive evidence behind their predictions, known as rationales. Rationale discovery is a widely adopted practice in the natural language processing (NLP) community \cite{DBLP:conf/kdd/Ribeiro0G16}, as well as in the graph \cite{DIR} and vision fields \cite{DBLP:conf/cvpr/ZhangYMW19}. Recently, this trend has also extended to the multi-modal community, with ImageQA-based tasks that request additional textual evidence \cite{DBLP:conf/cvpr/ParkHARSDR18, DBLP:conf/cvpr/DuaKB21} and causalvid that applies the concept to videoQA. Despite the progress, existing solutions often require a rationale finder with significant computation overhead \cite{IGV} or need to be trained in a contrastive manner \cite{EIGV}. RaFormer, however, can identify rationales (\ie the critical frames) without requiring additional training loss.

%% file: sec/2_preliminaries.tex
\section{Preliminaries}
\label{sec:preliminaries}
Taking a holistic view of existing methods, we summarize the paradigm of VideoQA and provide a formal definition in this section. Throughout the paper, we use upper-cased (\eg $A$) and lower-cased (\eg $a$) letters to denote a variable and its deterministic value, respectively.

% \vspace{5pt}
\noindent \textbf{Modeling.}
Given the video $V$ and the question $Q$, the VideoQA model $G$ aims to exploit information in both visual and textual streams to yield the predictive answer $\hat{A}$.
% \begin{gather}\label{eq:conventional-modeling}\
%     \hat{A} = f_{\hat{A}}(V,Q),
% \end{gather}
%
In leading VideoQA methods, $G$ is a combination of a video encoder $G_{V}$, a question encoder $G_{Q}$ and a fuser $G_{F}$.
% , \ie $f=f_{\hat{}} \,\circ f_{\hat{A}}$, where
% $f_{\hat{M}}:V,Q\to \hat{M}$ 
It aims to encapsulate the visual content and linguistic semantics into multi-modal
knowledge and then generate the predictive answer $\hat{A}$.
Typically, an entropy-based risk function is adopted to abridge the predictive answer $\hat{A}$ and the ground-truth answer ${A}$:
\begin{gather}\label{equ:erm-loss}
    \min\mathcal{L}((G_{V}(V), G_{Q}(Q) )\circ G_{F},\, A).
\end{gather}

% \vspace{5pt}
\noindent \textbf{Data representation.}
We take a video of $T$ clips and select the frame in the middle of each clip to serve as a representation of the entire video. Each of these frames is presented by extracting a frame feature $\vb{f}_t$ through a pretrained image recognition backbone, as well as $S$ object features $\vb{o}_{t,s}$ using a pretrained object detector, where $t$ and $s$ are the indices for frame and object, respectively. To represent the text, we use a pretrained language model as question encoder, which takes in the question as a sequence of $L$ tokens, and produces a textual representation $\vb{q}_l$ for each token. During training, the visual backbones are fixed while the language backbone is fine-tuned end-to-end, following the approach described in \cite{VGT}.
To create a common $d$-dimensional space for the representations, we apply three linear mappings to $\vb{f}t$, $\vb{o}_{t,s}$, and $\vb{q}_l$, respectively, resulting in $\vb{F}\!=\!\left\{ \vb{f}_t \right\}_{t=1}^{T} \!\in\!\vb{R}^{T\times d}$, 
$\vb{O}\!=\!\left\{ \vb{o}_{t,s}\right\}_{t=1,s=1}^{T,S}\!\in\!\vb{R}^{T\times S \times d}$, 
and 
$\vb{Q}\!=\!\left\{ \vb{q}_l \right\}_{l=1}^{L}\!\in\!\vb{R}^{L\times d}$, which represent the features of the frame, object, and question, respectively.

%% file: sec/3_method.tex
\section{Methodology}

\begin{figure}[t!]
\centering
\includegraphics[width=.47\textwidth]{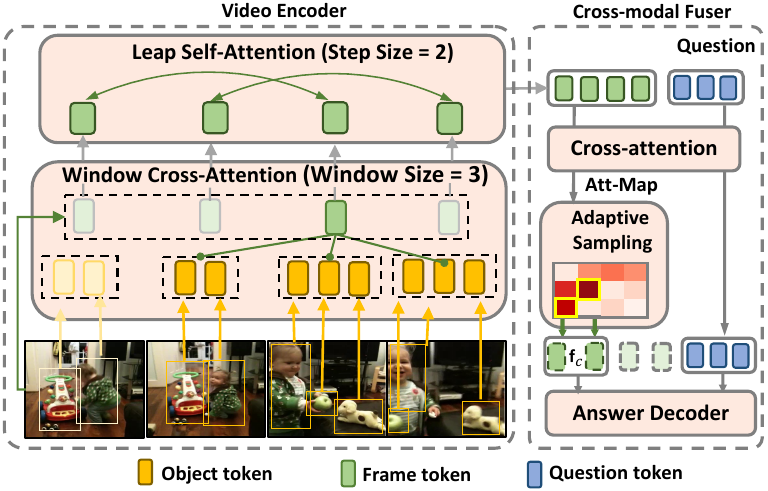}
\caption{Overview of RaFormer. We made the advancement on two modules: the video encoder and the cross-modal fuser. The video encoder incorporates a window-based cross-attention that aggregates objects within a temporal window to enhance the representation for each frame. On top of that, it also adopts a leap self-attention mechanism that avoid neighboring-frame redundancy by imposing attention only to distant frames. In the Cross-modal Fuser, a cross-attention takes the video and question tokens as input, yielding multi-modal knowledge and a cross-modal attention map. Based on this attention map, an adaptive sampling module identifies critical frame representations, which are then fed into a answer decoder for prediction.}
\vspace{-10pt}
\label{fig:main}
\end{figure}

In the existing literature, the temporal redundancy in video-language tasks has been analyzed by \cite{atp} from the dataset construction angle. However, a comprehensive solution for VideoQA has been missing until now. To fill this gap, we propose a novel VideoQA framework named Redundancy-aware Transformer (RaFormer). As depicted in \cref{fig:main}, RaFormer consists of a novel video encoder that incorporates a window cross-attention and a leap self-attention to mitigate the neighboring-frame redundancy. Additionally, it tackles the cross-modal redundancy by introducing an adaptive sampling module to cross-modal fuser, which discover the critical frames from a vision-language interaction perspective.

% \subsection{Overview}
\subsection{Video Encoder}
% 弄成transformer encoder 模块，现在的写法没有ffn
To encode a video instance, our video encoder aims to generate contextualized frame representations that focus on the detailed changes of objects within neighboring frames while avoiding neighboring-frame redundancy. Specifically, unlike previous encoder designs that resort to multi-level hierarchy \cite{hqga,hostr,VGT}, which introduces neighboring-frame redundancy that can overwhelm the object details, RaFormer focuses on the detail movement of objects within neighboring-frame by adopting a window cross-attention. It enhances the representation of center frames by aggregating the objects in its temporal neighborhood. On top that, we leverage a leap attention \cite{leapattention} to avoid imposing attention on neighboring frames, resulting a concise representation for input video.

\subsubsection{Window Cross-Attention}
Window Cross-Attention (WCA) aims to model the detailed movement of objects within a temporal window $W$, which helps to enhance a frame representation $\vb{f}_t \in \mathbb{R}^{d}$ by aggregating all objects within its temporal window. Concretely, we first add spatio-temporal position encoding to each object feature following \cite{VGT}, and then apply an intra-frame self-attention to contextualize all objects within a single frame $\vb{o}_t\!\in\!\mathbb{R}^{ S \times d}$. Based on the output tokens $\vb{o'}_{t}$, we flatten the objects within a temporal window $W$ (denote as $\vb{o}_t^W \!=\!\left\{ \vb{o}_{t,s}\right\}_{t-(W+1)/2,s=1}^{t+(W-1)/2,S}\!\in\!\mathbb{R}^{W\times S \times d}$) to the shape of $(WS) \times d$ and feed it as query to window cross-attention transformer decoder, along with the corresponding frame embedding $\vb{f}_t$ as key and value:
\begin{gather}
    \vb{o'}_{t}=\text{Self-Attention}(\vb{o}_{t})+\vb{o}_{t},\\
    \vb{f'}_{t}=\text{Cross-Attention}_{W}(\vb{o}_{t}^W)+\vb{o}_{t}^W, \quad \text{s.t.}\,|W|=w
\end{gather}
where $\text{Cross-Attention}_{W}$ denotes window cross-attention with window size $W$. In practice, we apply window cross-attention in multi-head manner, where each head is set to different window size. We empirically show that such multi-scale temporal reception fields are more flexible to cater different video content, and thus bringing in extra benefits. 

\subsubsection{Leap Attention}
To avoid neighbor-frame redundancy and preserve object details obtained in window cross-attention, we resort to leap attention \cite{leapattention} that computes self-attention on discrete temporal frame pairs. With this dilated strategy, we could build connection only between the distant frames, instead of recklessly merging all neighboring frame together, thus maintaining the detailed visual clue at object-level. 
Here, we use $E$ to denotes the temporal skipped step size. For example in \cref{fig:main}, when $E$=2, we get attention connection on following frame pairs for a sequence of four frames: (0,2), (1,3). 
Formally, by applying leap attention on the enhanced frame representations $\vb{f'}$, we acquire the final output of video encoder $\vb{f''}\in \mathbb{R}^{T \times d}$ as:
\begin{gather}
    \vb{f''}=\text{Leap-Attention}_E(\vb{f'})+\vb{f'}, \quad \text{s.t.}\,|E|=e, 
\end{gather}
where $\text{Leap-Attention}_E$ refers leap attention with step size of $E$. For simplicity, we omit the superscript and use $\vb{f}\in \mathbb{R}^{T \times d}$ to denote the output of video encoder.

\subsection{Cross-modal Fuser with Adaptive Sampling} \label{sec:rationlizer}
\begin{figure}[t!]
\centering
\includegraphics[width=0.47\textwidth]{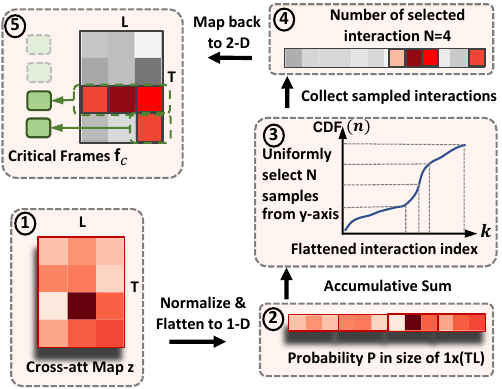}
\caption{Illustration of Adaptive Sampling process.}
\vspace{-10pt}
\label{fig:3}
\end{figure}

Although we have alleviated neighboring-frame redundancy with the video encoder design, cross-modal redundancy can still overwhelm the critical information when fusing frame and question tokens. To address that, we introduce Adaptive Sampling (AS), a parameter-free module that adaptively down-samples frame tokens according to their interaction activeness with question embedding. As shown in \cref{fig:3}, we first use the attention map between frame and question tokens to indicate their interaction activeness. Then, we flatten and normalize all $T \times L$ interactions to get a probability distribution $p$ over all interactions. Next, we calculate cumulative distribution function (CDF) based on $p$, and then select a subset of interactions using inverse transform sampling. Finally, we softly down-sample the frames by selecting tokens that corresponds to the sampled interactions, which adaptively remove redundant frames with the minimal computational overhead. 

% Aiming to identify the critical interactions from the multi-modal representations, rationalizer $f_{\hat{R}}$ should naturally coordinate the encoder structure with minimal computational overhead.
% %
% \lyc{Following this essence, $f_{\hat{R}}$ firstly establishes an interaction score that leverages the encoder's attention map with careful modification, then based on the interaction score, it collects rationales from the multi-modal representation in an interaction-aware manner.}

\subsubsection{Interaction Scoring}\label{sec:scoring}
To gather the critical frames, a naive solution is to generate an importance vector in size of $1\times T$ by prepending a  $\left\langle \text{CLS} \right\rangle$ token to the question, which denotes the contribution of each of frame tokens, and then we can collect the critical frames by selecting from this importance vector.
However, the collection under such a scheme suffers from \emph{uni-modal bias} \cite{rubi}. As a result, tokens with similar background can exert a high attention score, and will be more likely to be selected, while ignoring the the question-critical ones.
Instead, we identifies the critical frames via the cross-modal interactions -- a mutual agreement of vision and language modalities.
As shown in the right part of \cref{fig:main}, first apply a cross-attention using encoded frames $\vb{f}$ as query and and question embedding $\vb{q}$ as key and value, which yields the frame token $\vb{\bar{f}}$ as well as the pre-normalized cross-attention map $\vb{z} \in \mathbb{R}^{T\times L}$ as output:
\begin{gather}\label{eq:cross-att}
    \vb{\bar{f}}, \vb{z} = \text{Cross-Attention}(\vb{f},\vb{q}) + \vb{f}.
\end{gather}
Based on the cross-attention map $\vb{z}$, we apply our adaptive sampling.

% \vspace{-10pt}
\subsubsection{Adaptive Frame Sampling}
Adaptive Sampling (AS) aims to select $N$ interactions from $\vb{z} \!\in\!\mathbb{R}^{T \times L}$ ($N \ll  T\times L$), then we can collect the corresponding 1-D tokens from the 2-D interactions view. Notably, by keeping only one frame when selected repetitively, we end up with an adaptive frame collection where only a small proportion of $\bar{f}$ are selected as critical frames $\vb{f}^c$  (\ie $\left| \vb{f}^c \right|<=N$).

Specifically, to select N interactions from $\vb{z}$, a naive approach is to select top-N with highest score. However, this deterministic approach does not perform well, because it discards all interactions with lower scores. Some of these interactions, however, can be useful particularly at the earlier stages when the features are less discriminative.
Therefore, we resort to inverse transform sampling to sample interactions in an differentiable manner.
As shown in \cref{fig:3}, we first flatten $\vb{z}$ into the shape of ${1 \times (TL)}$, and then apply a softmax over all interactions to obtain the significance scores that can be interpreted as probabilities $\vb{p}\in\mathbb{R}^{1 \times (TL)}$:
\begin{gather}\label{eq:probility}
    \vb{p}=\text{Softmax}(\text{Flatten}(\vb{z})).
\end{gather}
% We first apply an element-wise softmax on $\vb{z}$ to obtain the significance scores $\vb{p}\!\in\!\mathbb{R}^{T \times L}$ of the interaction.
% \begin{gather}\label{eq:probility}
%     \vb{p}_{t,l}=\frac{\exp(\vb{z}_{t,l})}{\sum_{t=1}^{T}\sum_{l=1}^{L}\exp(\vb{z}_{t,l})}
% \end{gather}interactions based on their significance scores $\vb{p}$. 
% Then, we flatten $\vb{p}$ to $\vb{p'} \in \mathbb{R}^{1 \times TL}$, which can be interpreted as probabilities, and 
Then we calculate the cumulative distribution function (CDF) of $\vb{p}$:
\begin{gather}
    \text{DCF}_{k} = \sum_{1}^{k}\vb{p}_{k}, \quad \text{s.t.}\,k=(t-1)\times L + l,
\end{gather}
where $k \in \left[ 1,2,...,TL \right]$ is the flatten index of the $\vb{p}$.

Next, we obtain the sampling function by taking the inverse of the CDF as:
% (\ie $\text{DCF}^{-1}(n)), n\in\left[ 0,1 \right]$.
\begin{gather}
    k=\text{DCF}^{-1}(n),\quad n\in\left[0,1 \right].
\end{gather}
In other words, the significance score $\vb{p}$ is used to calculate the mapping function between the indices of the original interactions and the sampled
interactions. To obtain $N$ samples, we use a fixed sampling strategy by choosing $n=\left\{ \frac{1}{2N},\frac{3}{2N},...\frac{2N-1}{2N} \right\}$. 
Since $\text{DCF}^{-1} \in \mathbb{R}$, we consider the indices of the tokens with the nearest significant scores as the sampling indices. 
% Notably, such sampling can benefit the training by maintaining the diversity of tokens at the early stages when the features are less discriminative.

After acquiring the $N$ sampled interactions, we collect their corresponding frame tokens as critical frames $\vb{f}_c$. It is worth noticing that gathering 1-D tokens from the 2-D interaction view can incur repetition. Thus, we only keep one frame if it is selected multiple times, which enables an adaptive collection of critical frames.

\begin{figure}[t!]
\centering
\includegraphics[width=0.47\textwidth]{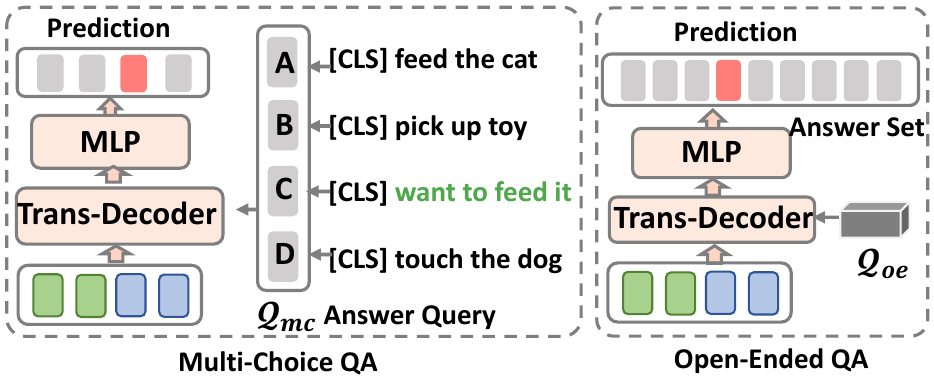}
\caption{Illustration of our decoding process in Multi-choice QA (left) and Open-end QA (right).}
\vspace{-10pt}
\label{fig:decoder}
\end{figure}

\subsection{Answer Prediction}
%general formulation of decoder
To generate the predicted answer, we use a transformer-based answer decoder that receives an answer query to extract the predictive information from the multi-modal features, and select the the predicted answer $\hat{A}$ from the answer candidates. According to task setting,  the answer candidates can be provided as $\left|A_{mc}\right|$ sentences that are tailored for each instance in the case of multi-choice QA, or as global answer categories of size $\left|A_{oe}\right|$ that are shared among all instances (\aka, Open-ended QA). This lead to the two formations of answer query (see in \cref{fig:decoder}).

\subsubsection{Multi-Choice QA} 
% In this prevailing setting, we first prepend a $\left\langle \text{CLS} \right\rangle$ token to each answer candidate to get their holistic representation, using the same language model for the question encoding. Then we gather them from $ \left|A_{mc}\right|$ answer candidates to form the answer query $\mathcal{Q}_{mc} \in \mathbb{R}^{\left|A_{mc}\right| \times d}$.  
%
In the Multi-Choice setting (see \cref{fig:decoder} left), the answer candidates are given as multiple sentences tailored for each instance. To generate a holistic representation for each candidate answer, we prepend a $\left\langle \text{CLS} \right\rangle$ token and use the same language model in the question encoder to process the each answer candidate in parallel. Then we gather the processed $\left\langle \text{CLS} \right\rangle$ tokens from $ \left|A_{mc}\right|$ answer candidates representations and form the answer query $\mathcal{Q}_{mc} \in \mathbb{R}^{\left|A_{mc}\right| \times d}$.
During decoding, we feed the answer query $\mathcal{Q}_{mc}$ as the query and the concatenation of critical frames and question (\ie $\left[ \vb{f}_c;\vb{q} \right]$) as key and value to the transformer decoder, and the decoder outputs the embedding $\vb{h}_{mc} \in \mathbb{R}^{\left|A_{mc}\right| \times d}$ as:
\begin{gather} \label{eq:decoder-mc}
    \vb{h}_{mc} = \text{Transformer-Decoder}(\mathcal{Q}_{mc}, \left[ \vb{f}_c;\vb{q} \right])
\end{gather}
where $\left[;\right]$ refers to the concatenation operation.
Notably, since the correctness of a answer candidate is invariant to its position, answer query is free of position encoding. 
Finally, we apply a linear projection on $\vb{h}_{mc}$ to obtain the predictive answer $\vb{\hat{a}_{mc}} \in \mathbb{R}^{\left|A_{mc}\right|}$. 
\begin{gather}
    \vb{\hat{a}}_{mc}=\text{MLP}(\vb{h}_{mc})
\end{gather}

% \vspace{-10pt}
\subsubsection{Open-ended QA}
In the open-ended setting, it is common practice \cite{hqga, IGV} to treat each candidate answer in the global answer set as a category, which makes the size of the answer set too large to be processed as multi-choice. Therefore, we initialize a single learnable embedding, denoted as $\mathcal{Q}_{oe} \in \mathbb{R}^{d}$, as the answer query for this setting.

Similar to \cref{eq:decoder-mc}, we use the transformer decoder to process the answer query $\mathcal{Q}{oe}$ along with the concatenation of critical frames and question ($[\vb{f}c;\vb{q}]$) as the key and value. This results in a decoded representation $h_{oe} \in \mathbb{R}^{d}$. Finally, we project $h_{oe}$ to the answer space $\mathbb{R}^{\left|A_{oe}\right|}$ to obtain the prediction $\vb{\hat{a}}_{oe}$.
\begin{gather}
    % \vspace{-2pt}
    \vb{\hat{a}}_{oe}=\text{MLP}(h_{oe}).
        % \vspace{-2pt}
\end{gather}

To align the predictive answer with the gold answer, we establish our objective on a cross-entropy loss.

%% file: sec/4_results.tex
\section{Experiments}
\input{tab/dataset}

\input{tab/causal_vid}

\input{tab/main_exp}

In this section, we show the experimental results that answer the following research questions:
\begin{itemize}[leftmargin=*]
\setlength\itemsep{-.20em}
    \item \textbf{RQ1:} How effective is RaFormer compared with the State-of-the-Art (SoTA) models?
    % \item \textbf{RQ1:} How does IGV perform against the State-of-the-Art (SoTA) methods and variation of backbone model?
    \item \textbf{RQ2:} How do modules of Raformer and parameter setting affect the performance?
    \item \textbf{RQ3:} What learning pattern does RaFormer capture?
    % \vspace{-15pt}
\end{itemize}
\noindent\textbf{Datasets:}
To evaluate the effectiveness of our proposed design, we conduct experiments on four benchmark datasets that challenge RaFormer from different aspects. Specifically, we use the prevalent Multi-Choice setting of \textbf{NExT-QA} \cite{next-qa} and \textbf{Causal-VidQA} \cite{causalvid} to test the model's temporal reasoning ability with complex causal and commonsense relations. As a complement, we also use the Open-Ended setting of \textbf{MSVD-QA} \cite{DBLP:conf/mm/XuZX0Z0Z17} and \textbf{MSRVTT-QA} \cite{DBLP:conf/mm/XuZX0Z0Z17}, which mainly emphasize the description of video objects, activities, and their attributes. For detailed statistics of the datasets, please refer to \cref{tab:dataset}.

% \vspace{5pt}
\noindent\textbf{Implementation Details:}
We followed the standard procedure established in \cite{VGT,IGV, EIGV} for sampling videos, where each video is represented as a sequence of $T=16$ frames, and each frame is encoded using a ViT-L \cite{dynamicViT} model pre-trained on ImageNet-21k. To extract object-level features, we used a Faster-RCNN \cite{faster-rcnn} model pre-trained on the Visual Genome, which detects $S=20$ objects on each frame. For encoding the textual input (questions and answers), we used a pre-trained Deberta-base model \cite{deberta} as the question encoder. During training, we optimized the model using an Adam optimizer with a learning rate of 1e-5, and set the hidden dimension to 768. For for all datasets, we set skip step-size $E$=4 in leap attention and sample $N=10$ interactions in adaptive sampling.

\subsection{Main Result (RQ1)}

\cref{tab:causal_vid} and \cref{tab:main} show that RaFormer's performance far exceeds existing SoTA methods. Specifically, our observations are as follows:
% 不同dataset的表现差异， 上更好，因为mc的视频和问题都更长，更难。
\begin{itemize}[leftmargin=*]
\setlength\itemsep{+.20em}
    % C 和basline 差不多
    \item \textbf{QA settings.}
    When comparing RaFormer's performance across four datasets, we observe that it provides a greater improvement to Multi-Choice QA than to Open-Ended QA. Specifically, NExT-QA and Causal-VidQA saw an increase of 3.9\% and 3.5\%, respectively, while MSVD-QA and MSRVTT only saw an increase of 2.6\% and 2.3\%, respectively, compared to previously reported results. This difference can be explained by two aspects:
    Firstly, Multi-Choice QA datasets tend to have longer videos with more objects, which inevitably introduce more neighboring-frame redundancy, compared to the Open-ended setting. Luckily, our video encoder is better suited to handle neighboring-frame redundancy, thus can bring more favorable improvement to Multi-Choice QA.
    % as window cross-attention is designed to model object-level changes in adjacent frames and the leap attention helps with neighboring-frame redundancy.
    Secondly, Multi-Choice datasets involve complex reasoning scenarios with composite question sentences and long videos, leading to more cross-modal redundancy. In contrast, Open-Ended datasets typically feature simpler questions and shorter videos. Therefore, our adaptive sampling strategy is particularly well-suited for this scenario, which   achieves a larger gain in Multi-Choice QA by wiping out more environmental frames.
    
    \item \textbf{Question type.} 
    Narrowing down the analysis to Multi-Choice datasets (NExT-QA and CausalVid-QA), it is apparent that the performance gain is primarily due to improvement in the composite question type (including: Acc@C\&Acc@T in NExT-QA, Acc@E\& Acc@P\& Acc@C in Causal-VidQA), which involves deeper understanding such as causal relation and counterfactual thinking. In contrast, the improvement in the descriptive question type is much more moderate (Acc@D: +0.5\% on NExT-QA, and +1.0\% on CausalVid). This highlights RaFormer's exceptional reasoning ability for complex questions.
    In particular, there is a significant improvement in reason-based questions on Causal-VidQA (Q$\to$AR: Acc@P +2.8\%, Acc@C +6.9\%), where the model has to justify its prediction by selecting the correct evidence. This naturally aligns with the mechanism of adaptive sampling, making RaFormer an ideal solution for this question type.
    
    \item \textbf{Benchmark size.}
    Regarding the performance improvement on Open-Ended datasets, it is interesting to note that the gain on MSRVTT-QA (+2.6\%) is larger than that on MSVD-QA (+2.3\%), despite sharing the exactly same question type. We suggest that the difference in dataset volume may have played a role in this observation, since MSVD-QA is five times smaller than MSRVTT-QA (\cf \cref{tab:dataset}). This could limit the reasoning ability of Raformer on MSVD-QA due to a smaller amount of training data. This highlights the potential of Raformer when trained on larger and more diverse datasets.
\end{itemize}

\vspace{-10pt}
\subsection{In-Depth Study (RQ2)}
\subsubsection{Ablative Results}
We scrutinize some key designs in RaFormer by performing model ablation and discussing other implementation alternatives.
The observations of \cref{tab:ablation} are as follows: 
\input{tab/ablation}

\begin{itemize}[leftmargin=*]
\setlength\itemsep{-.20em}
    \item In the top section, we study the design of our video encoder.  In the 1st row, we study the effectiveness of window cross-attention by replacing it with a conventional object merging method \cite{VGT, hqga} (\ie apply mean pooling to merge all objects in a single frame, then add the pooled representation to the frame feature), where the performance drop of this alternative (``w/o WCA'') demonstrates the necessity of capturing the object transition details in adjacent frames.
    In 2nd row, we verify the efficacy of Leap-Attention (LA) in reducing the neighboring-frame redundancy by replacing it with a vanilla self-attention. As a result, we also witness a performance drop on this variant (``w/o LA''), which demonstrates cutting off connections between adjacent frames and restricts them to the distant frames can remove the redundancy from temporal dimension.
    In the 3rd row, we ablate the whole encoder design by removing both window attention and leap attention, the denote this variant as ``w/o WCA \& LA''. We notice the performance of this variant drops more significantly, which means the benefits of WCA and LA focus on different aspect of video encoding, (WCA for modelling object detail in adjacent frames, LA help to restrict neighboring-frame redundancy) and their contribution can be added together.
    
    \vspace{5pt}
    \item In the bottom section, some mutants of the adaptive sampling are investigated. First, we study the impact of our sampling strategy by removing the adaptive sampling module (``w/o AS''), where all frames are used for answer prediction without differentiating their interactions with question. As expected, the accuracy drop demonstrates the negative impact of in redundant interactions in answer reasoning as well as the necessity of critical frame down-sampling.
    Next, we show that collecting critical interactions via hard ranking of (\ie Hard TopN) interaction scores results in sub-optimal performance.
    The reason is that, in hard topN, interaction with a relatively smaller score will not be explored, even if the scores itself are far from discriminative at the early stage, thus leading to poor optimization. However, RaFormer circumvents this issue by sampling from inversed cumulative distribution function, which enable a better exploration of the interactions in a differentiable manner.
    As discussed in \cref{sec:scoring}, we also analyze the benefits of interaction-aware sampling by comparing to an alternative that collects critical frames from a probability vector generated $\left\langle \text{CLS} \right\rangle$ token. As expected, this variant causes a sharp performance decline. We empirically find that this is because of the \emph{uni-modal bias} issue mentioned in \cref{sec:scoring}, \ie, the high correlated adjacent frames can induce a selection bias that prompts visually similar tokens as critical frames, instead of the answer-responsive ones. 

    \vspace{5pt}
    \item Finally, we show the effectiveness of our overall design by ablating both encoder (implemented as ``w/o WCA \& LA'') and adaptive sampling (implemented as ``w/o AS''). We show this variant, denoted as ``w/o Enc \& LA'', performs poorly comparing to RaFormer, which testifies that our overall design is reasonable.
    
\end{itemize}

% \begin{figure}[t!]
% \centering
% \includegraphics[width=0.47\textwidth]{fig/ablation.pdf}
% \caption{(a) Study of (}
% \vspace{-5pt}
% \label{fig:ablation}
% \end{figure}

% \vspace{-15pt}

\subsubsection{Study of Window Size}
In order to investigate the effect of window cross-attention, the performance of RaFormer is studied using different window sizes $W$. Specifically, the default setting uses 4-head attention, where different heads can have the same window size or varied ones. As shown in \cref{tab:step}, when setting all heads' window size to 1, the model delivers the worst accuracy, which demonstrates the necessity of modeling the objects in the neighboring frame. As the window size increases, the model delivers better results on MSRVTT when the window size is equal to 3, while NExT-QA reaches its peak at $W$=5. This discrepancy may be caused by the complexity of the video source. Videos in NExT-QA feature relational reasoning among multiple objects, where a larger window size can incorporate more temporal related objects, thus benefiting the relational reasoning. In comparison, MSRVTT typically questions the attributes of a single object, making a smaller window size sufficient to capture the answer information.
Finally, the results in last row show that none of the single-size window settings can beat the multi-scale encoding, where four attention heads are equipped with different window sizes. This demonstrates that multi-scale window settings are more flexible in catering to different video content.

\begin{figure*}[t]
  \centering
  \includegraphics[width=0.95\textwidth]{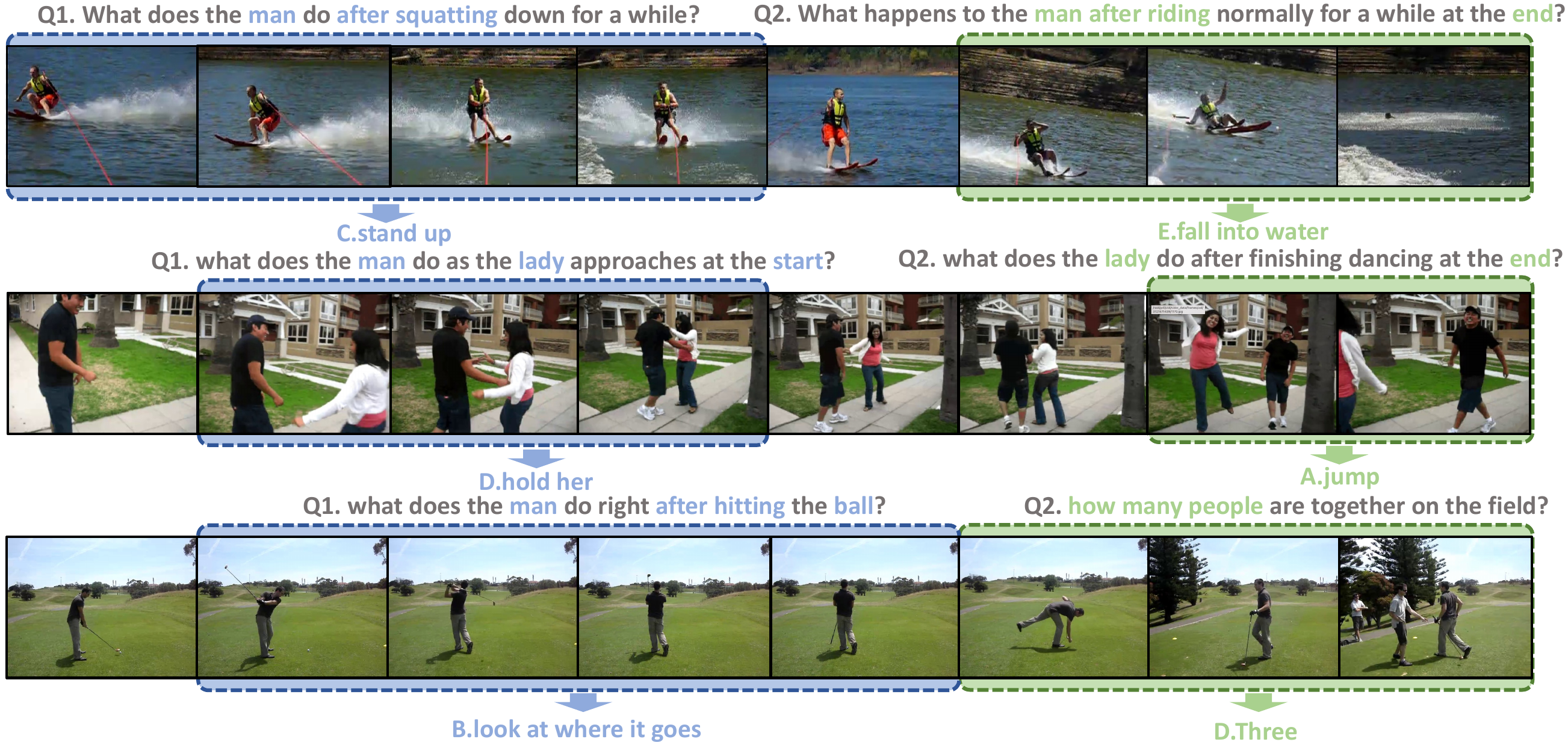}
  	% \vspace{-8pt}
 \vspace{-0.15in}
  \caption{Case-Study on NExT-QA. Each video comes with two questions that focus on different parts of the video. The \blue{blue} and \textcolor[rgb]{0,0.5,0}{green} windows indicate the selected critical frames. In addition, we also mark the question tokens that correspond to the sampled interactions.}
   	\vspace{-0.15in}
  \label{fig:case-study}
 	% \vspace{-13pt}
\end{figure*}

\subsubsection{Study of Sample Size N}
\begin{figure}[t!]
    \centering
    \begin{minipage}{.5\linewidth}
        \begin{minipage}{1\textwidth}
        \input{tab/step_size}
        \captionof{table}{WCA's window size W.}
        \label{tab:step}
        \end{minipage} 
        \begin{minipage}{1\textwidth}
        % \begin{subfigure}[t]{1\linewidth}
            \includegraphics[width=\textwidth]{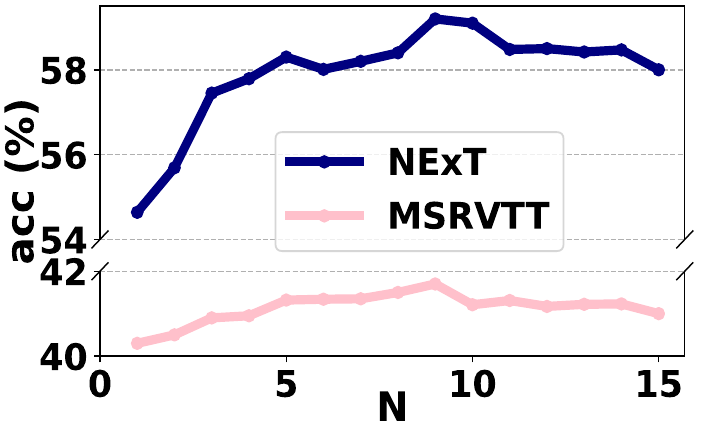}
            \vspace{-20pt}
            \caption{Hyper-parameter N in adaptive sampling.}
            \label{fig:topN}
        \end{minipage} 
    \end{minipage}
    \hspace{5pt}
    \begin{minipage}{.43\linewidth}
        % \begin{subfigure}[t]{1\linewidth}
            \includegraphics[width=1.1\textwidth, height = 1.4\textwidth]{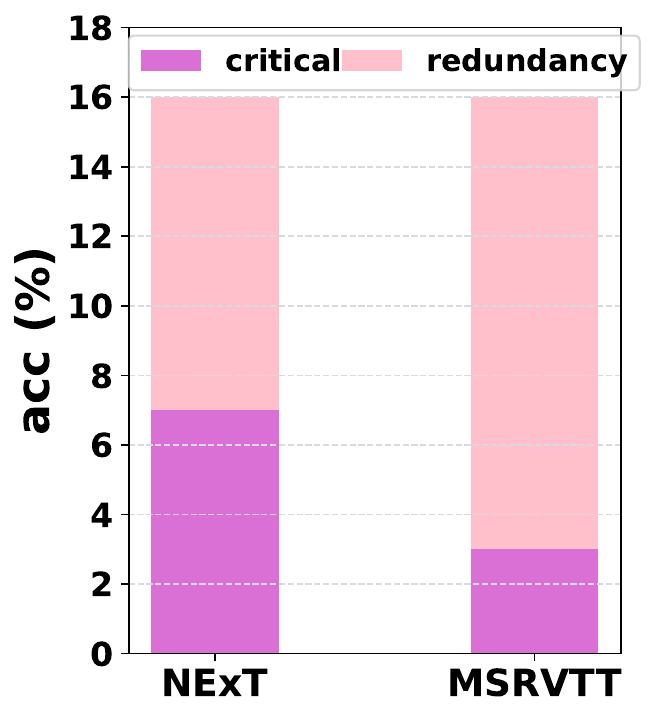}
            \vspace{-15pt}
            \caption{The number of the critical frames and removed tokens after adaptive sampling.}
            \label{fig:rationale}
        % \end{subfigure}
    \end{minipage}
    % \vspace{-10pt}
    % \caption{(a)Study of hyper-parameter N. (b) Feature analysis. (c) The number of the discovered rationale tokens and removed redundancy tokens in Video and Question. }
    % \vspace{-15pt}
     \vspace{-0.15in}
\end{figure}

To validate how the size of the sampled interaction $N$ affect RaFormer, we conduct experiments with a variation of $N$ and show the results in \cref{fig:topN}. In both datasets, we observe constant performance gains as $N$ increases from 1, and the peaks are reached at around 10. Then, as $N$ keeps growing, the accuracies in both datasets drop since more redundancies are introduced. 
It is worth noticing that, RaFormer can surpass all existing methods even with $N$ equal to 1 (\ie only one frame is select and used to infer the answer), which shed light on the redundancy nature of VideoQA task, and the deficiency of currents methods as their ignorance of this inherent redundancy.
Compare to MSRVTT-QA, fluctuation on NExT-QA is more drastic when $N$ is small, where the altering can cause a 5\% difference in accuracy. 
This is because questions of MSRVTT-QA generally focus on the description of a single entity, where the crtical frames are more perceivable even with small $N$. In contrast, the NExT-QA typically questions the temporal relation in a composite scenario, where a small $N$ is unable to cover the golden frames that delivers the answer information, thus leading to sub-optimal performance.

% \vspace{-5pt}
\subsubsection{Study of Critical Frames}
To grasp the learning insight of RaFormer, we analyzed the number of identified critical frames in the original videos and present the statistics in \cref{fig:rationale}. As expected, our adaptive sampling can identify a small group of tokens as critical frames while leaving the rest as redundant. Specifically, we observed that NExT-QA requires more frames to answer questions than MSRVTT-QA. This is due to the relational reasoning nature of NExT-QA, which requires clues from multiple entities, and are typically associated with more frames. In contrast, MSRVTT-QA mainly focuses on the description of a single object, which can be encoded with fewer frames. Interestingly, although RaFormer's peak performance on MSRVTT-QA and NExT-QA is achieved with almost the same size of critical interaction group (both around $N$=10 \cf \cref{fig:topN}), the number of critical frames differs significantly between the two datasets. This demonstrates that our frame selection strategy can adaptively cater to different VideoQA instances.

\subsection{Qualitative analysis (RQ3)}
Intuitively, the design of adaptive sampling naturally empowers RaFormer with visual explainability. Following this line, \cref{fig:case-study} shows some predictive results on NExT-QA along with the identified critical frames. To better understand learning pattern of adaptive sampling, we also mark the question tokens that corresponds to the sampled interactions, which can be interpreted as "critical words". Here, each video instance comes with two questions that inquire about visual clues at different frames.
Generally, RaFormer can locate very few indicative elements that involve the interaction of high activitness. For the critical words, RaFormer tends to imply the entity in the video (\eg ``man", ``lady") as well as words that serve as a temporal indicator (\eg ``after", ``end"). As for the identified critical frames, we notice that our adaptive sampling tends to select frames that contains the key entities in the question (\eg ``man", ``lady"). Similarly, it also select frames that corresponds to the temporal indicator, which benefits critical frames selection by narrowing down the target scope. This is attributed to our cross-modal sampling strategy, where the answer is referred to as a mutual agreement between critical frames and question words.
Apart from that, we also notice that, even for the same video, RaFormer can accredit different scenes as the critical frames in response to the question clues. Such adaptiveness demonstrates the design philosopy of our sampling module, where frames are differentiate base on their interaction activeness with question, making different question target at different scene.

%% file: tab/dataset.tex
\setlength{\tabcolsep}{3pt}
\begin{table}[t!]
  \centering
  \small
  \caption{Dataset statistics. MC: Multi-Choice OE: Open-Ended}\scalebox{0.97}{
    \begin{tabular}{lccccc}
    \toprule
    \toprule
    Dataset & Challenge & \#QA pair  & V-Len & Q-Len & QA \\
    \midrule
    NExT-QA & Causal \& Temporal & 48K   & 44s    &   11.6    & MC \\
    Causal-VidQA & Evidence \& Commonsense & 161K  & 9s     &   9.5    & MC \\
    MSVD-QA  & Description &   50K    & 10s    &    6.6   & OE \\
    MSRVTT-QA & Description & 244K  & 15s    &   7.4    & OE \\
    \bottomrule
    \end{tabular}}%
  \label{tab:dataset}%
  \vspace{-7pt}
\end{table}%

%% file: tab/causal_vid.tex
\setlength{\tabcolsep}{10pt}
\begin{table*}[t]
    \small
    \centering
    \caption{Accuracy (\%) comparison on Causal-VidQA. D: Description, E: Explanation, P: Prediction, C: Counterfactual. *: Reproduced result using official implementation.
    }
     \vspace{-0.15in}
    \scalebox{0.95}{
    \begin{tabular}{l|c|cccccccc|c}
    \toprule
    \toprule
    \multirow{3}*{Methods} & 
    \multirow{3}*{Venue} &
    \multicolumn{9}{c}{Causal-VidQA} \\ 
    \cline{3-11}
    ~ & ~  & \multirow{2}*{Acc@D} & \multirow{2}*{Acc@E} & \multicolumn{3}{c}{Acc@P} & \multicolumn{3}{c|}{Acc@C} & \multirow{2}*{Acc@All} \\
    \cline{5-10}
    ~ & ~ &~ & ~ & Q $\rightarrow$ A & Q $\rightarrow$ R & Q $\rightarrow$ AR & Q $\rightarrow$ A & Q $\rightarrow$ R & Q $\rightarrow$ AR &~ \\ 
    \midrule
    HCRN \cite{hcrn} & CVPR20 & 56.4 & 61.6 & \underline{51.7} & 51.3 & \underline{32.6} & 51.6 & 53.4 & 32.7 & 48.1 \\
    HGA \cite{hga} & AAAI20 & 65.7 & \underline{63.5} & 49.4 & 50.6 & 32.2 & 52.4 & 55.9 & 34.3 & 48.9 \\
    B2A \cite{park2021bridge} & CVPR21 & 
    \underline{66.2} & 62.9 & 49.0 & 50.2 & 31.2 & \underline{53.3} & \underline{56.3} & \underline{35.2} & \underline{49.1} \\
    VGT* \cite{xiao2020visual} & ECCV22 & \bf{70.8} & \bf{70.3} & \bf{55.2} & \bf{56.9} & \bf{38.4} & {\bf61.0} & \bf{59.3} & \bf{42.0} & \bf{55.4}\\
    \midrule
    RaFormer (Ours) & - & \bf{71.8} & \bf{73.8} & \bf{58.5} & \bf{58.5} & \bf{41.2} & {\bf67.1} & \bf{64.1} & \bf{48.9} & \bf{58.9} \\
     Abs. Improve & - & +1.0 & +3.5 & +2.7 & +1.6 & +2.8 & +6.1 & +4.8 & +6.9 & +3.5\\
    \bottomrule
    \end{tabular}
    }
    \label{tab:causal_vid}
    % \vspace{-0.4cm}
\end{table*}

%% file: tab/main_exp.tex
\setlength{\tabcolsep}{1pt}
\begin{table}[t!] 
  \centering
  \small
  \caption{Accuracy (\%) comparison on NExT-QA, MSVD-QA, and MSRVTT-QA. VIB: Visual Inductive Bias, G,H denotes graph- and hierarchy- based method, respectively. ROI: whether use additional region feature as additional input. Acc@C, T, D, denote questions type of Causal, Temporal, and Descriptive in NExT-QA, respectively. The \textbf{best} and \underline{2nd best} results are highlighted.}
  \scalebox{1}{
    \begin{tabular}{l|c|ccc|c|c|c}
    \toprule
    \toprule
    \multirow{2}*{Methods} & 
    \multirow{2}*{venue}&
    \multicolumn{4}{c|}{NExT-QA} & 
    \multirow{2}*{MSVD} & \multirow{2}*{MSRVTT} \\
     \cline{3-6} 
    ~ & ~ & Acc@C & Acc@T & Acc@D & Acc@All & ~ & ~ \\
    
%     \multicolumn{2}{c|}{Encoder} & \multicolumn{4}{c|}{NExT-QA}  & \multirow{2}[4]{*}{MSVD-QA} & \multirow{2}[4]{*}{MSRVTT-QA} \\
% \cmidrule{2-7}          & IB & ROI & Acc@C & Acc@T & Acc@D & Acc@All &       &  \\

    \midrule
    HCRN \cite{hcrn}  & CVPR20)     & 47.1  & 49.3  & 54.0    & 48.9  & 36.1  & 35.6 \\
    HGA \cite{hga}   & AAAI20     & 48.1  & 49.1  & 57.8  & 50.0    & 34.7  & 35.5 \\
    HOSTR \cite{hostr} & IJCAI21     & -     & -     & -     &  -    & 39.4  & 35.9 \\
    PGAT \cite{pgat}  & MM21     & -     & -     & -     &  -    & 39.0    & 38.1 \\
    MHN \cite{MHN}   & IJCAI22    & -     & -     & -     &  -    & 40.4  & 38.6 \\
    IGV \cite{IGV}   & CVPR22     & 48.6  & 51.7  & 59.6  & 51.3  & 40.8  & 38.3 \\
    HQGA \cite{hqga}  & AAAI22      & 49.0    & 52.3  & 59.4  & 51.8  & 41.2   & 38.6 \\
    EIGV \cite{EIGV}  & MM22   & 51.2  & 51.5  & 61.0    & 52.9  & \underline{43.7}  & 39.3 \\
    VGT \cite{VGT}   & ECCV22     &51.6  & 51.9  & 63.7  & 53.7  & -  & \underline{39.7} \\
    VGT-PT \cite{VGT} & ECCV22 &  \underline{52.8} &  \underline{54.5} &  \underline{67.3} &  \underline{55.7} & - & - \\
    \midrule
    RaFormer &   -     & \bf{58.2}  & \bf{57.7}    & \bf{67.8}  & \bf{59.6}  & \bf{46.0}  & \bf{42.3} \\ 
    Abs.Improve &  -  &  +5.4   & +3.2  & +0.5  & +3.9 & +2.3  & +2.6 \\
    \bottomrule
    \end{tabular}}
  \label{tab:main}%
  \vspace{-15pt}
\end{table}%

%% file: tab/ablation.tex
\setlength{\tabcolsep}{4pt}
\begin{table}[t!] 
  \small
  \centering
  \caption{Ablative study on NExT-QA and MSRVTT-QA.}
 \vspace{-0.15in}
  \scalebox{1}{
    \begin{tabular}{l|ccc|c|c}
    \toprule
    \toprule
    \multirow{2}*{Variants} & 
    \multicolumn{4}{c|}{NExT-QA} & 
    \multirow{2}*{MSRVTT} \\
    \cline{2-5} 
    ~ & Acc@C & Acc@T & Acc@D & Acc@All & ~ \\
    
%     \multicolumn{2}{c|}{Encoder} & \multicolumn{4}{c|}{NExT-QA}  & \multirow{2}[4]{*}{MSVD-QA} & \multirow{2}[4]{*}{MSRVTT-QA} \\
% \cmidrule{2-7}          & IB & ROI & Acc@C & Acc@T & Acc@D & Acc@All &       &  \\
    \midrule
    w/o WCA & 57.7 & 57.3 &  67.9 & 59.2 & 41.1 \\
    w/o LA  & 58.4 & 57.0 & 65.7 & 59.2 & 41.3 \\
    w/o  WCA \& LA &  57.1 & 56.3 & 65.6 & 58.2 & 40.8 \\
    \midrule
    % w/o col-softmax & - & - & - & - \\
    w/o AS &  57.1 &  56.6 &  67.3 & 58.7 & 41.4 \\
    % w/o $\norm{\vb{v}_t}$ & 57.8 & 57.1 & 64.8 & 58.7 & 41.0 \\
    Hard TopN & 56.9 & 57.6 & 68.3 & 59.0 & 41.7 \\
    $\left\langle \text{CLS} \right\rangle$ Sampling & 57.0 & 56.2 & 65.0 & 58.1 & 41.3 \\
    \midrule
    w/o Enc \& AS & 57.3 & 55.4 & 64.1 & 57.9 & 40.3 \\
    \bf{RaFormer} & \bf{58.2}  & \bf{57.7}  & \bf{67.8}  & \bf{59.6}  & \bf{42.3} \\
    \bottomrule
    \end{tabular}}
    \vspace{-7pt}
  \label{tab:ablation}%
\end{table}%

%% file: tab/step_size.tex
% % Table generated by Excel2LaTeX from sheet 'Sheet1'
% \setlength{\tabcolsep}{1pt}
% % \begin{table}[t!] 
%   \small
%   \centering
%   \caption{Ablative study on NExT-QA and MSRVTT-QA.}
%   % \vspace{-0.3cm}
%   \scalebox{0.7}{
%     \begin{tabular}{l|ccc|c|c}
%     \toprule
%     \toprule
%     \multirow{2}*{Variants} & 
%     \multicolumn{4}{c|}{NExT-QA} & 
%     \multirow{2}*{MSRVTT} \\
%     \cline{2-5} 
%     ~ & Acc@C & Acc@T & Acc@D & Acc@All & ~ \\
    
% %     \multicolumn{2}{c|}{Encoder} & \multicolumn{4}{c|}{NExT-QA}  & \multirow{2}[4]{*}{MSVD-QA} & \multirow{2}[4]{*}{MSRVTT-QA} \\
% % \cmidrule{2-7}          & IB & ROI & Acc@C & Acc@T & Acc@D & Acc@All &       &  \\
%     \midrule
%     1 & 49.8 & 50.7 & 50.2 & 50.1 & 38.1 \\
%     3 & 49.8 & 50.7 & 50.2 & 50.1 & 38.1 \\
%     5 & 49.8 & 50.7 & 50.2 & 50.1 & 38.1 \\
%     7 & 49.8 & 50.7 & 50.2 & 50.1 & 38.1 \\

%     \midrule
%     $\left[ 1,3,5,7 \right]$ & \bf{58.1}  & \bf{57.6}  & \bf{65.6}  & \bf{59.2}  & \bf{41.5} \\
%     \bottomrule
%     \end{tabular}}
%     % \vspace{-10pt}
%   \label{tab:step}%
% % \end{table}%

\setlength{\tabcolsep}{3pt}
% \begin{table}[t!] 
  \small
  \centering
  % \caption{Test on window size $W$.}
  % \vspace{-0.3cm}
      \scalebox{0.9}{
    \begin{tabular}{c|cc}
    \toprule
    \toprule
    $W$ & NExT-QA & MSRVTT \\
    
%     \multicolumn{2}{c|}{Encoder} & \multicolumn{4}{c|}{NExT-QA}  & \multirow{2}[4]{*}{MSVD-QA} & \multirow{2}[4]{*}{MSRVTT-QA} \\
% \cmidrule{2-7}          & IB & ROI & Acc@C & Acc@T & Acc@D & Acc@All &       &  \\
    \midrule
    1 & 58.7 & 41.7 \\
    3 & 59.2 & \underline{42.0} \\
    5 & \underline{59.4} & 41.8 \\
    7 & 58.9 & 41.9 \\

    \midrule
    $\left[ 1,3,5,7 \right]$ & \bf{59.6}  & \bf{42.3} \\
    \bottomrule
    \end{tabular}}
    % \vspace{-10pt}
  % \label{tab:step}%
% \end{table}%

%% file: sec/6_conclusions.tex
% \vspace{-5pt}
\section{Conclusions}
% \vspace{5pt}
In this paper, we pinpoint the redundancy issue in current VideoQA paradigm. Specifically, we design a novel video encoder to emphasize the modelling of detailed object movement within the neighboring frames, while address the neighboring-frame redundancy by imposing the leap attention that models the frame-level representations in a dilate manner. To tackle the cross-model redundancy in prevailing fuser design, we incorporate an adaptive sampling strategy that select a small set of critical frames according to their interactions with question tokens. Extensive experiments on four benchmark datasets have demonstrated the superior of RaFormer. We hope this simple yet effective design can spark more future efforts in handling VideoQA redundancy.